\documentclass[10pt,twocolumn,letterpaper]{article}

\usepackage{cvpr}              

\usepackage{amsmath,amsfonts,bm}

\usepackage{natbib}
\usepackage{url}
\usepackage{enumitem}
\usepackage{float}
\usepackage{multirow}
\usepackage{textcomp}
\usepackage{tabularx}
\usepackage[normalem]{ulem}
\usepackage{footnote}
\usepackage{threeparttable}
\usepackage{caption}
\usepackage[utf8]{inputenc}
\usepackage{amssymb}             
\usepackage{mathrsfs} 
\usepackage{CJKutf8}
\usepackage[table]{xcolor}
\usepackage{indentfirst}
\usepackage{threeparttable}
\usepackage{array}
\usepackage{stfloats}
\usepackage{verbatim}
\usepackage{colortbl}
\usepackage{makecell}
\usepackage{rotating}
\usepackage{arydshln}
\usepackage{wrapfig}
\usepackage{pifont}

\usepackage{algorithm}
\usepackage{algpseudocode}

\usepackage[table]{xcolor} 
\usepackage{ulem} 
\usepackage{tikz} 

\definecolor{mlb}{RGB}{173,216,230}  
\definecolor{mlo}{RGB}{255,223,186}  

\definecolor{cvprblue}{rgb}{0.21,0.49,0.74}
\usepackage[pagebackref,breaklinks,colorlinks,allcolors=cvprblue]{hyperref}

\def\paperID{7586} 
\def\confName{CVPR}
\def\confYear{2026}

\title{CineSRD: Leveraging Visual, Acoustic, and Linguistic Cues for Open-World Visual Media Speaker Diarization}

\author{
\textbf{Liangbin Huang}$^{1,\dagger}$, \textbf{Xiaohua Liao}$^{1,\dagger}$, \textbf{Chaoqun Cui}$^{2,3,\dagger}$, \textbf{Shijing Wang}$^{4}$, \textbf{Zhaolong Huang}$^{1}$,\\ \textbf{Yanlong Du}$^{1}$, \textbf{Wenji Mao}$^{2,3,*}$\\
$^1$Hujing Digital Media and Entertainment Group\\
$^2$MAIS, Institute of Automation, Chinese Academy of Sciences\\
$^3$School of Artificial Intelligence, University of Chinese Academy of Sciences\\
$^4$School of Computer Science and Technology, Beijing Jiaotong University\\
\texttt{huangliangbin.hlb@alibaba-inc.com, wenji.mao@ia.ac.cn}
}

\begin{document}
\maketitle

\begingroup
\renewcommand\thefootnote{} 
\footnotetext{$^\dagger$ Equal contribution.}
\footnotetext{$^*$ Corresponding author.}
\endgroup

\begin{abstract}

Traditional speaker diarization systems have primarily focused on constrained scenarios such as meetings and interviews, where the number of speakers is limited and acoustic conditions are relatively clean. To explore open-world speaker diarization, we extend this task to the visual media domain, encompassing complex audiovisual programs such as films and TV series. This new setting introduces several challenges, including long-form video understanding, a large number of speakers, cross-modal asynchrony between audio and visual cues, and uncontrolled in-the-wild variability. To address these challenges, we propose Cinematic Speaker Registration \& Diarization (CineSRD), a unified multimodal framework that leverages visual, acoustic, and linguistic cues from video, speech, and subtitles for speaker annotation. CineSRD first performs visual anchor clustering to register initial speakers and then integrates an audio language model for speaker turn detection, refining annotations and supplementing unregistered off-screen speakers. Furthermore, we construct and release a dedicated speaker diarization benchmark for visual media that includes Chinese and English programs. Experimental results demonstrate that CineSRD achieves superior performance on the proposed benchmark and competitive results on conventional datasets, validating its robustness and generalizability in open-world visual media settings.

\end{abstract}    
\section{Introduction}

In multi-party dialogue scenes involving multiple speakers, a crucial challenge is to accurately identify and assign segments of speech or transcribed text to their corresponding speakers, a task known as speaker diarization. Speaker diarization aims to integrate multimodal machine learning techniques by combining visual, acoustic, and linguistic cues to determine “who spoke when” in video or audio streams \cite{app5,intro1,app6,app3,app4,intro2}. It plays a significant role in applications such as meeting transcription, human-computer interaction, forensic audio analysis, and video dubbing/subtitling \cite{app1,app2,vd1,vd2,vd3}. Early speaker diarization methods mainly relied on audio-based unimodal features \cite{modal1,app8,modal3}. To mitigate the effects of noise and overlapping speech, researchers have proposed bimodal audio-visual \cite{modal4,modal5,modal6} and audio-textual \cite{modal7,modal8,modal9} approaches. More recent studies explore methods that jointly fuse audio, visual, and semantic information to leverage cross-modal complementarity for improving the robustness and interpretability of speaker diarization. However, most of these methods focus on constrained scenarios such as meetings and interviews, characterized by a small number of speakers, simple scenes, and distinctive features. With the rise of online video platforms and the film and television industry in recent years, the application contexts of speaker diarization have become more complex and dynamic, exposing the growing limitations of existing methods \cite{vd2,app9}.

In this study, we extend the speaker diarization task to the visual media domain, aiming to explore more generalizable and robust approaches for open-world scenarios. The goal of visual media speaker diarization is to associate each line in audiovisual programs, such as films and TV series, with its corresponding character (speaker). Visual media programs typically cover diverse genres and themes, making speaker diarization on such videos significantly more complex than in constrained scenarios like meetings and interviews, thereby constituting an open-world setting. Specifically, visual media speaker diarization faces several major challenges: 1) \textbf{Long-form video understanding} - audiovisual programs often have lengthy durations, with films typically lasting around two hours and TV series accumulating dozens of hours across episodes; 2) \textbf{Large number of speakers} - a single audiovisual program may include dozens of characters, far exceeding the number of speakers in meetings or interviews scenarios; 3) \textbf{Audio-visual asynchrony} - in audiovisual programs, a character’s voice does not necessarily coincide with their face being visible, posing challenges for aligning audio and visual features; 4) \textbf{In-the-wild nature} - visual media are filmed in open, real-world environments, exhibiting uncontrolled variations such as diverse acoustic conditions and complex visual dynamics.

To address the above challenges, we propose \textbf{Cine}matic \textbf{S}peaker \textbf{R}egistration \& \textbf{D}iarization (CineSRD), a unified multimodal speaker diarization framework designed for open-world visual media scenarios. CineSRD leverages visual, acoustic, and linguistic cues from the video, speech, and subtitles of audiovisual programs to achieve speaker annotation. Specifically, since audiovisual programs typically lack prior knowledge of the number of speakers, CineSRD first performs a \textit{Visual Anchor Clustering} process for speaker registration, obtaining an initial speaker set. It then employs an Audio Language Model (ALM) to integrate speech and subtitle information, together with speech timbre features, to conduct the \textit{Speaker Turn Detection} process. Based on the detected turns, CineSRD further conducts \textit{Off-Screen Speaker Supplementation} to refine speaker annotation and recover unregistered off-screen speakers. In addition, we construct and release a dedicated speaker diarization benchmark for visual media scenarios, which includes three subsets (Chinese, Chinese-Hard for dialect programs, and English) to evaluate speaker annotation accuracy across different language audiovisual programs. 

In summary, our main contributions are as follows:
\begin{itemize}
\item We extend the speaker diarization task to the open-world setting of visual media and present a complete solution along with a benchmark for performance evaluation.
\item We propose CineSRD, a training-free speaker diarization framework that leverages multimodal cues to achieve speaker annotation for visual media programs.
\item We construct and release a benchmark for evaluating visual media speaker diarization, which supports bilingual (Chinese and English) and dialectal acoustic conditions.
\item Experimental results demonstrate that CineSRD not only achieves superior performance on our benchmark but also performs competitively on conventional benchmarks.
\end{itemize}
\section{Related Work}



\subsection{Application Scenarios of Speaker Diarization}

Speaker diarization has long served as a foundational component in multi-speaker audio understanding tasks, supporting downstream applications such as automatic meeting transcription, multimedia retrieval, and human-computer interaction \cite{app1,app2}. In conversational settings, identifying “who spoke when” enhances both the structural and semantic organization of dialogues, providing critical information for ASR, emotion recognition, and conversational analysis \cite{app3,app4}. Traditional systems were primarily developed for constrained acoustic environments such as teleconferences or interviews \cite{app5,app6}. However, with the rise of open-domain media (e.g., audiovisual programs and online videos), systems must handle complex situations involving a larger number of speakers, speech overlaps, off-screen speakers, and noise interference \cite{app7,app8,app9}. Therefore, in this study, we focus on visual media, which encompass programs of various genres, representing an in-the-wild scenario that requires methods capable of performing speaker registration and recognition across diverse video scenes.

\subsection{Evolution of Modalities in Speaker Diarization}

The evolution of speaker diarization methods mirrors the progression from unimodal to multimodal learning. Early diarization systems relied solely on acoustic cues, which suffered from noise and overlapping speech \cite{modal1,app8,modal3}. To overcome these limitations, researchers began integrating additional modalities. Audio-visual approaches exploited lip motion and facial synchronization to disambiguate overlapping voices \cite{modal4,modal5,modal6,hermes}. Subsequently, audio-textual methods leveraged semantic cues from transcribed speech to detect speaker turns and role-specific linguistic patterns \cite{modal7,modal8,modal9}. Recent studies further integrate audio, visual, and semantic information within a unified optimization framework, demonstrating that modality complementarity can substantially enhance the robustness and interpretability of diarization \cite{modal10,modal11,aclsd}. Although these studies still focus on relatively simple scenarios such as meeting transcription, this trend marks a shift in speaker diarization from unimodal signal-driven processing toward multimodal cognitive reasoning. Following the recent study \cite{hermes}, we propose the CineSRD framework, which performs speaker diarization in complex visual media scenes by leveraging multimodal information from audiovisual programs with modern multimodal recognition and ALM technologies.

\section{CineSRD Framework}
\label{sec:allmethod}

\subsection{Task Definition and Notations}

The visual media speaker diarization task aims to automatically annotate the speakers of lines in audiovisual programs of genres such as films and TV series by leveraging information from the video, speech, and text modalities. Specifically, for an audiovisual program $\mathcal{P}$ containing multiple lines $s$, each line can be associated with its corresponding video segment $\text{v}(s)$ and audio (speech) segment $\text{a}(s)$ according to the timing notes in the program’s subtitles (see example subtitle format in the Supplementary Material), i.e., $\mathcal{P}\equiv \{(s,\text{v}(s),\text{a}(s))\}$. Meanwhile, the program contains multiple speakers (characters), represented as a set $\mathcal{R}$. CineSRD aims to model and learn the many-to-one mapping from program lines to speakers, denoted as $\Phi:\mathcal{P}\rightarrow\mathcal{R}$. The preprocessing procedure of CineSRD is presented in detail in the Supplementary Material.

\subsection{Overall Framework}

\begin{figure*}[t]
  \centering
  \includegraphics[width=\textwidth]{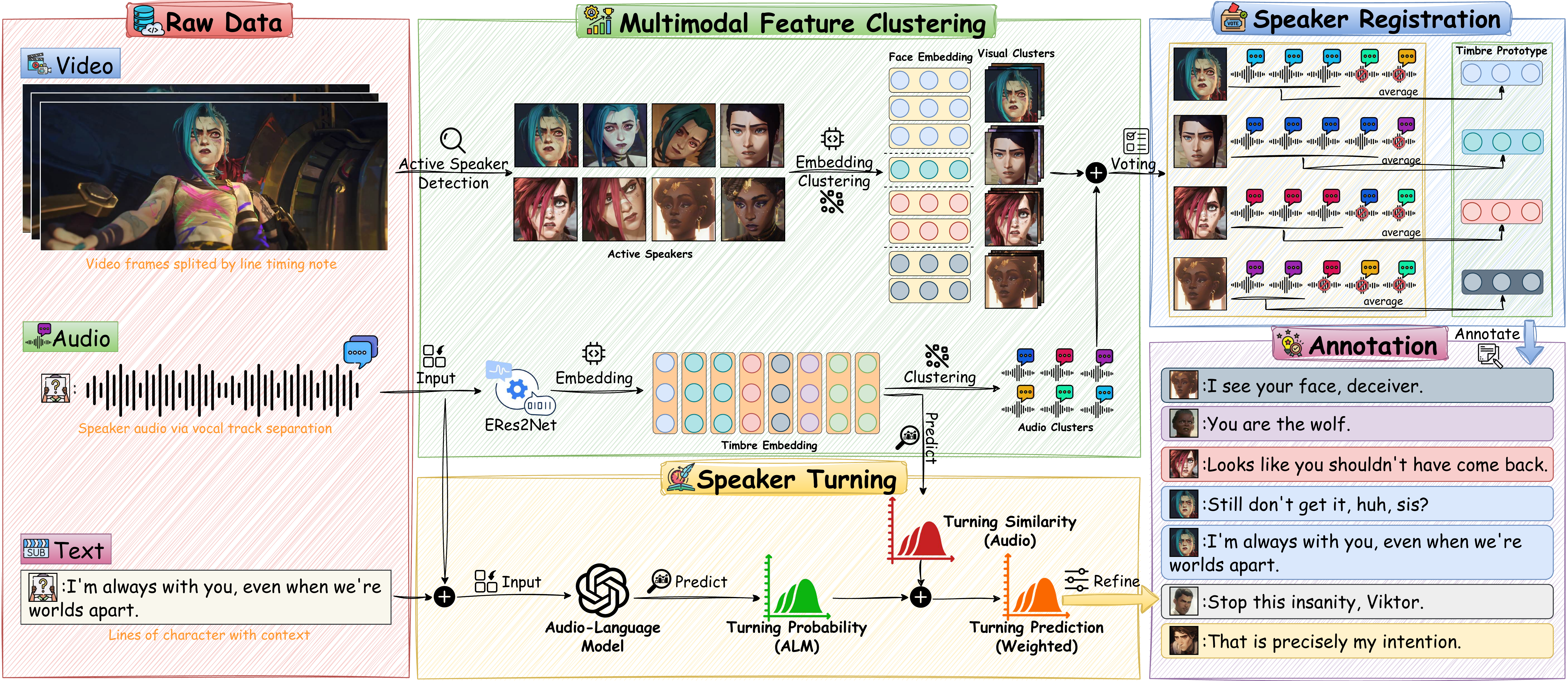}
  \caption{The overall structure of the unified framework CineSRD for visual media speaker diarization.}
  \label{fig:framework}
\end{figure*}

We present the overall design of the CineSRD framework in Figure~\ref{fig:framework}. CineSRD leverages multimodal information to perform speaker diarization in visual media scenes. We first apply face and timbre related techniques to cluster and align the video and speech modalities, thereby enabling speaker registration within the program, obtaining timbre prototype embeddings for multiple speakers, and achieving initial speaker labeling. Then, we employ an ALM to combine speech and text modalities for speaker turn detection, allowing the supplementation of new speakers and refinement of speaker annotations. We next detail the method design of CineSRD.

\subsection{Visual Anchor Clustering}

For each line $s \in \mathcal{P}$, the speaker’s speech is generally played within its corresponding time span. However, the visible face in the video may not always belong to the speaker (audio-visual asynchrony), resulting in off-screen speakers. Therefore, for the visual modality, we employ an active speaker detection model to determine whether the video segment $\text{v}(s)$ corresponding to line $s$ contains an active speaker. Lines without detected active speakers are reserved for later processing. For lines with detected active speakers, we use a face embedding model to extract their face embeddings $f_{\text{v}}(s)$ and perform clustering (e.g., spectral clustering~\cite{sc}) to assign each line a visual cluster label $c_{\text{v}}(s)\in \left \{1,\cdots ,n_{\text{v}}\right \}$, where $n_{\text{v}}$ denotes the number of visual clusters. For the audio modality, we use a timbre embedding model to extract the timbre embedding $f_{\text{a}}(s)$ of each line’s audio segment $\text{a}(s)$. Similarly, we cluster all audio embeddings and assign each line an audio cluster label $c_{\text{a}}(s)\in \left \{1,\cdots ,n_{\text{a}}\right \}$, where $n_{\text{a}}$ denotes the number of audio clusters.

Since the facial features extracted from the visual modality are generally more discriminative, the visual clustering results are typically more reliable than those from the audio modality. Therefore, we use the visual clusters as anchors for speaker registration, as illustrated in Figure~\ref{fig:register}. Specifically, each line $s \in \mathcal{P}$ has its corresponding face embedding $f_{\text{v}}(s)$ (if an active speaker is detected), timbre embedding $f_{\text{a}}(s)$, visual cluster label $c_{\text{v}}(s)$, and audio cluster label $c_{\text{a}}(s)$. For each visual cluster $i$, we collect the set of lines it contains, $\mathcal{S}_{i}=\{s\in\mathcal{P}\mid c_{\text{v}}(s)=i\}$, and treat this cluster as a registered speaker. Within $\mathcal{S}_{i}$, we perform a voting process over audio clusters:
\begin{equation}
\hat{c}_{\text{a}}(i) = \underset{k}{\arg\max} \big|\{s \in \mathcal{S}_i \mid c_{\text{a}}(s) = k\} \big|
\end{equation}
The audio cluster $\hat{c}_{\text{a}}(i)$ with the highest vote count is regarded as the audio cluster corresponding to speaker $i$. The mean timbre embedding of the lines in this cluster is used as the timbre prototype embedding of the registered speaker:
\begin{equation}
\mu_i=\frac{1}{|\mathcal{T}_i|}\sum_{s\in \mathcal{T}_i} f_{\text{a}}(s),\quad \mathcal{T}_i=\{s\in \mathcal{S}_{i}\mid c_{\text{a}}(s)=\hat{c}_{\text{a}}(i)\}.
\end{equation}
Finally, each visual cluster $i$ is registered as a speaker, and the entire set of speakers is represented as $\mathcal{R}=\{(i,\mu _{i})\mid i=1,\cdots ,n_{\text{v}}\}$.

\begin{figure}[!h]
  \centering
  \includegraphics[width=0.4\textwidth]{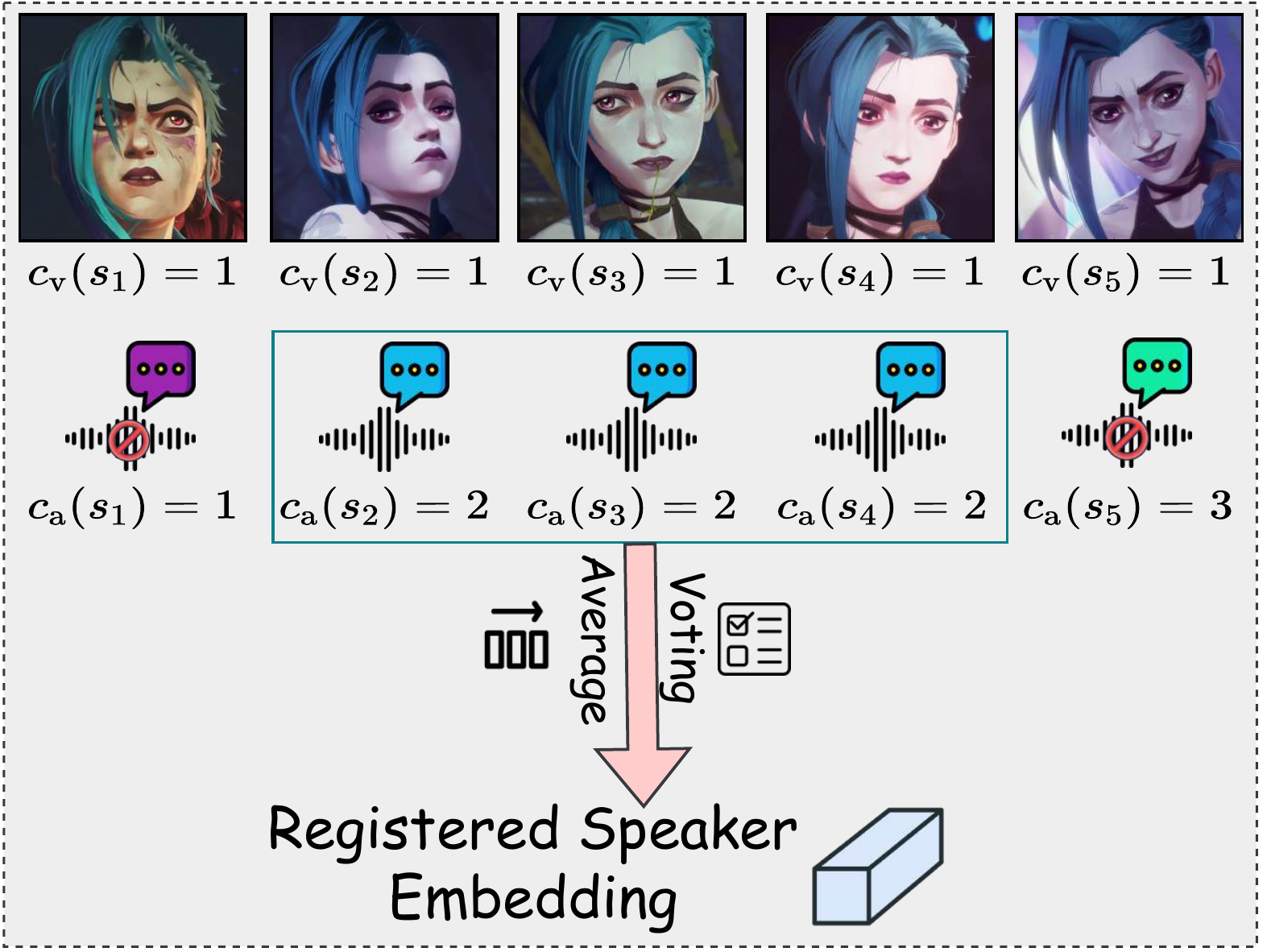}
  \caption{Demonstration of speaker registration.}
  \label{fig:register}
\end{figure}

\subsection{Speaker Turn Detection}

\begin{table*}[t]
\centering
\caption{The new speaker supplementation process. \textit{Similarity} represents the timbre embedding similarity of adjacent lines.}
\resizebox{\textwidth}{!}{
\begin{tabular}{clcccccc}
 \Xhline{1.0pt}
 \rowcolor{gray!20}
 \textbf{Group} & \textbf{Line} & \textbf{Similarity} & \textbf{Active} & $\sigma(s)$ & $\sigma(G)$ & \textbf{Operation} \\
 \hline
 \multirow{3}{*}{1} & \texttt{You've got a good heart.} & - & \ding{51} & 1.0 & \multirow{3}{*}{1.0} & \multirow{3}{*}{-}\\
 ~ & \texttt{Don't ever lose it.} & 0.651 & \ding{51} & 1.0 & ~ & ~ \\
 ~ & \texttt{No matter what happens.} & 0.712 & \ding{51} & 1.0 & ~ & ~ \\
 \hdashline
 \multirow{2}{*}{2} & \texttt{Power, real power,} & 0.198 & \ding{55} & 0.185 & \multirow{2}{*}{0.210} & \multirow{2}{*}{Register new speaker}\\
 ~ & \texttt{doesn't come to those who were born strongest.} & 0.684 & \ding{55} & 0.235 & ~ & ~ \\ 
 \hdashline
 3 & \texttt{I thought maybe you could love me like you used to.} & 0.155 & \ding{51} & 1.0 & 1.0 & - \\
 \hdashline
 \multirow{3}{*}{4} & \texttt{In the pursuit of great,} & 0.210 & \ding{55} & 0.312 & \multirow{3}{*}{0.285} & \multirow{3}{*}{\makecell[l]{Register new speaker or merge\\with existing new speakers}}\\
 ~ & \texttt{we failed to do good.} & 0.695 & \ding{55} & 0.264 & ~ & ~ \\
 ~ & \texttt{It's time to make things right.} & 0.742 & \ding{55} & 0.279 & ~ & ~ \\
 \Xhline{1.0pt}
\end{tabular}
}
\label{tab:newspeaker}
\end{table*}

The above initial speaker registration process relies on facial and timbre features from the video and audio modalities while ignoring the textual semantics of lines. This leads to several issues: (1) due to audio-visual asynchrony, using visual clusters as anchors for speaker registration cannot cover all lines and characters, resulting in missing off-screen speakers, especially minor or supporting roles; (2) the audio modality may suffer from recognition errors caused by background noise, overlapping speech, or similar vocal timbres among characters. In contrast, the text modality provides semantic cues that enable coherent contextual reasoning and accurate identification of dialogue boundaries. Therefore, we incorporate textual information. Specifically, we perform speaker turn detection by integrating audio and text modalities, followed by new speaker supplementation and correction of the initial diarization results.

We first employ an ALM that integrates audio and textual semantics to determine whether two adjacent lines are spoken by the same speaker. To ensure sufficient contextual information for each adjacent line pair, we input multiple consecutive lines (10 in our experiments) along with their corresponding audio into the ALM. For each adjacent line pair, the ALM generates a label token such as “0” or “1” (“1” indicating the same speaker), with generation probabilities denoted as $p_0$ and $p_1$, respectively. The prediction probability of the ALM for a pair is thus defined as $P_{\text{alm}}=p_1/(p_0+p_1)$. See the Supplementary Material for the prompt and response formats of the ALM. In addition, we compute the cosine similarity of the timbre embeddings for each adjacent line pair and normalize it, denoted as $S_{\text{tim}}$. The final prediction of whether two adjacent lines are spoken by the same speaker is obtained by a weighted combination of the ALM prediction probability $P_{\text{alm}}$ and the timbre similarity $S_{\text{tim}}$:
\begin{equation}
P_{\text{std}}=w\cdot P_{\text{alm}}+(1-w)\cdot S_{\text{tim}},
\end{equation}
where $w$ is a tunable hyperparameter that balances the contributions of audio and text.

\subsection{Off-Screen Speaker Supplementation}
\label{sec:spk_sup}

Due to the audio-visual asynchrony problem in the visual modality, active speaker detection cannot cover all lines in $\mathcal{P}$. For lines where no active speaker is detected, their timbre embeddings are compared with all speaker prototype embeddings $\mu_i$ in $\mathcal{R}$ using cosine similarity, and the speaker label is assigned based on the highest similarity. Since some undetected off-screen speakers may still exist among these lines, we design a new speaker supplementation strategy.

Specifically, based on the results of speaker turn detection, we take adjacent line pairs identified as different speakers ($P_{\text{std}}<0.5$) as group boundaries. In this way, all lines within each group $G$ belong to the same speaker. Since speaker turn detection combines textual semantics and audio features for prediction, it achieves high accuracy, and we therefore regard its results as reliable. A main speaker is determined for each group through a voting mechanism, and all lines in the group are standardized to this speaker. Then, for each line $s\in G$, we determine a new speaker score $\sigma(s)$ based on whether an active speaker is detected:
\begin{equation}
\begin{aligned}
\sigma(s) = \mathbb{I}(s) + \left(1 - \mathbb{I}(s)\right) \max_{1 \le i \le n_{\text{v}}} \text{sim}(f_{\text{a}}(s), \mu_i),
\end{aligned}
\end{equation}
where $\mathbb{I}(s)$ is an indicator function, which equals 1 when an active speaker is detected for line $s$, and 0 otherwise. Given the high accuracy of face-related techniques, we assume that when an active speaker is detected from the video corresponding to line $s$, the speaker annotation is reliable, and thus set $\sigma(s)=1$. Next, we compute the average new speaker score for group $G$ as:
\begin{equation}
\sigma(G)=\frac{1}{|G|}\sum_{s\in G} \sigma(s).
\end{equation}
If $\sigma(G)$ falls below the threshold $\eta$, the average timbre embedding of the lines in $G$, denoted as $\mu_G=\sum_{s\in G} f_{\text{a}}(s)/|G|$, is registered as a new speaker. The lines within this group are then labeled as belonging to this new speaker. If a new speaker has already been registered, it is merged with existing ones based on a timbre similarity threshold $\epsilon$. An illustration of the new speaker supplementation process is provided in Table~\ref{tab:newspeaker}.

\subsection{SubtitleSD Benchmark}

To quantitatively evaluate the performance of CineSRD, we construct and release a visual media speaker diarization benchmark, named SubtitleSD. We build SubtitleSD using subtitles, speech, and video from film and television programs on the online video platform Youku. For each line in the subtitles of these programs, we manually annotate the corresponding speaker (character). The programs span multiple themes, including romance, fantasy, comedy, suspense, and sitcom. We divide the programs into three subsets according to language and difficulty: Chinese, Chinese-Hard, and English. The Chinese-Hard subset includes one program that features multiple Chinese dialects and a very large number of speakers (317). Due to its challenging acoustic conditions and complex speaker distribution, we use it to evaluate visual media speaker diarization methods under extreme conditions. Statistics of the SubtitleSD benchmark are shown in Table~\ref{tab:sta}. A demo of SubtitleSD data is provided in the Supplementary Material.

\begin{table}[!h]
\centering
\caption{Statistics of SubtitleSD dataset.}
\resizebox{0.48\textwidth}{!}{
\begin{tabular}{cccc}
\Xhline{1.0pt}
\rowcolor{gray!20}
\textbf{Statistic} & \textbf{Chinese} & \textbf{Chinese-Hard} & \textbf{English} \\
\hline
\textbf{language} & CN & CN (dialect) & EN \\
\textbf{\# duration (h)} & 70.98 & 16.96 & 4.65 \\ 
\textbf{\# avg speaker} & 127.83 & 317 & 19.5 \\ 
\textbf{\# lines} & 80752 & 15528 & 5227 \\ 
\textbf{\# avg line token} & 6.14 & 6.92 & 9.05 \\ 
\Xhline{1.0pt}
\end{tabular}
}
\label{tab:sta}
\end{table}

We compare SubtitleSD with existing speaker diarization datasets/benchmarks in Table~\ref{tab:comp}. Unlike traditional speaker diarization datasets that primarily focus on relatively simple scenarios such as meetings, interviews, and debates, SubtitleSD targets the complex, in-the-wild domain of visual media. Visual media videos are typically long-form; therefore, although SubtitleSD contains fewer videos than several existing datasets, its total video duration exceeds most of them. Moreover, the average number of speakers per video is significantly higher than in other datasets, highlighting the high difficulty of the visual media speaker diarization task and imposing stronger requirements on the robustness of different methods in open-world scenarios.

\begin{table*}[t]
\centering
\caption{Comparisons with existing speaker diarization datasets/benchmarks. ASV denotes the average number of speakers per video. TDS denotes the total number of distinct speakers.}
\begin{tabular}{lcccccc}
\Xhline{1.0pt}
\rowcolor{gray!20}
\textbf{Dataset} & \textbf{Modality} & \textbf{Scenario} & \textbf{\# Videos} & \textbf{Total Duration} & \textbf{ASV} & \textbf{TDS}\\
\hline
AMI Corpus~\cite{AMI} & AV & meetings & 684 & 100h & 4.0 & 189 \\
AVDIAR~\cite{modal5} & AV & chat & 27 & 21m & 2.2 & 11\\
VoxConverse~\cite{VoxConverse} & A & debate, news & 448 & 63h50m & 5.6 & - \\
AVA-AVD~\cite{modal6} & AV & documentaries, moviesbie & 351 & 29h15m & 7.7 & 1500 \\
\hdashline
SubtitleSD (Ours) & AVT & multi-genre visual media & 130 & 92h35m & 21.2 & 1054 \\
\Xhline{1.0pt}
\end{tabular}
\label{tab:comp}
\end{table*}

\section{Experiments}

\subsection{Experimental Settings}

We evaluate the performance of CineSRD and other baseline methods on the visual media speaker diarization task using our SubtitleSD benchmark. In addition, we conduct experiments on AVA-AVD, a traditional speaker diarization benchmark, to verify the generalization ability of SubtitleSD in other scenarios. Since AVA-AVD is an audio-visual benchmark, we eliminate the speaker turn detection process in CineSRD that relies on the text modality, and perform off-screen speaker supplementation using only timbre similarity $S_{\text{tim}}$. We compare our approach with the following baseline methods:
\begin{itemize}
\item \textbf{Spectral Clustering (SC)} \cite{sc} constructs an affinity graph based on feature similarity and performs speaker clustering via spectral decomposition.
\item \textbf{Agglomerative Hierarchical Clustering (AHC)} \cite{ahc} iteratively merges speech segments with the highest cosine similarity to achieve multi-speaker clustering.
\item \textbf{AVR-Net} \cite{modal6} introduces a learnable modality mask to dynamically adjust feature weights based on whether the speaker appears on screen.
\item \textbf{EC2P} \cite{aclsd} propagates constraints across audio, visual, and semantic modalities to optimize the similarity matrix and improve clustering performance.
\end{itemize}
We adopt consistent irrelevant parameters across the main experiments and ablation studies to ensure fair and coherent comparisons, and all important settings are listed in Table~\ref{tab:hp}. The source code for CineSRD and the SubtitleSD benchmark are available at \url{https://github.com/BSTLL/CineSRD}.

\begin{table*}[h]
\centering
\caption{Hyperparameter configuration in experiments.}
\resizebox{\textwidth}{!}{
\begin{tabular}{cccl}
 \Xhline{1.0pt}
 \rowcolor{gray!20}
 \textbf{Type} & \textbf{Setting} & \textbf{Value} & \textbf{Remark} \\
 \hline
 \multirow{6}{*}{\textbf{Models}} & active speaker detection & TalkNet \cite{talknet} & - \\
 ~ & face detection & RetinaFace \cite{retinaface} & - \\
 ~ & face embedding & CurricularFace \cite{curricular} & - \\
 ~ & face quality assessment & FQA & from Alibaba DAMO Academy in ModelScope \\
 ~ & timbre embedding & ERes2NetV2 \cite{eres2netv2} & - \\
 ~ & ALM & Qwen2-Audio-7B \cite{Qwen2-Audio} & the ALM used in speaker turn detection \\
 \hline
 \multirow{5}{*}{\textbf{Hyperparameter}} & temperature & 1.2 & \multirow{3}{*}{ALM generation} \\
 ~ & top k & 50 & ~ \\
 ~ & top p & 0.95 & ~ \\
 \cdashline{2-4}
 ~ & $w$ & 0.45 & speaker turn detection weight \\
 ~ & $\eta$ & 0.45 & off-screen speaker supplementation threshold \\
 \Xhline{1.0pt}
\end{tabular}
}
\label{tab:hp}
\end{table*}

\subsection{Results and Discussion}

\subsubsection{Visual Media Speaker Diarization}

We present the evaluation results on our SubtitleSD benchmark in Table~\ref{tab:sdeval-main}, including performance on the Chinese, Chinese-Hard, and English subsets. We report Diarization Error Rate (DER) \cite{der} and Jaccard Error Rate (JER) \cite{jer}, and additionally report the performance of EC2P and our CineSRD when the text modality is eliminated. The results show that CineSRD achieves a clear improvement over other baseline methods, and even surpasses EC2P under its tri-modal setting while CineSRD uses only dual modalities. This demonstrates the importance of using the visual modality as an anchor during clustering: face features offer higher discriminability, and thus the visual modality should serve as the reference when fusing audio-visual clustering outputs. Moreover, the results indicate that pure audio clustering methods such as AHC and SC are significantly limited under multilingual and complex conditions, struggling with long-form video processing, audio-visual asynchrony, and large numbers of speakers. This highlights the necessity of multimodal feature fusion in speaker diarization. Leveraging mechanisms such as visual anchor clustering, CineSRD achieves strong performance even in dialect-heavy scenarios like Chinese-Hard, validating its high generalization ability and robustness in open-world programs.

\begin{table*}[t]
\centering
\caption{Speaker diarization evaluation in open-world visual media scenario. CineSRD adopts Spectral Clustering to perform audio-visual modality clustering. The 1st and 2nd best results are denoted as \colorbox{mlb}{\textbf{blue}} and \colorbox{mlo}{\textbf{orange}}, respectively.}
\begin{tabular}{lcccc|ccc|ccc}
\Xhline{1.0pt}
\rowcolor{gray!20}
~ & ~ & \multicolumn{2}{c}{\textbf{Chinese}} & & \multicolumn{2}{c}{\textbf{Chinese-Hard}} & & \multicolumn{2}{c}{\textbf{English}} & \\
\cline{3-4} \cline{6-7} \cline{9-10}
\rowcolor{gray!20}
\multirow{-2}{*}{\textbf{Method}} & \multirow{-2}{*}{\textbf{Modality}} & \textbf{DER~$\downarrow$} & \textbf{JER~$\downarrow$} & & \textbf{DER~$\downarrow$} & \textbf{JER~$\downarrow$} & & \textbf{DER~$\downarrow$} & \textbf{JER~$\downarrow$} & \\
\hline
AHC~\cite{ahc} & A & 0.13982 & 0.45223 & & 0.20679 & 0.57714 & & 0.12481 & 0.41017 & \\
SC~\cite{sc} & A & 0.14945 & 0.47114 & & 0.22255 & 0.65194 & & 0.20116 & 0.43402 & \\
\hdashline
\multirow{2}{*}{EC2P \cite{aclsd}}  & AV & 0.13595 & 0.41524 & & 0.21125 & 0.58984 & & 0.11951 & 0.39785 & \\
~ & AVT & 0.13451 & 0.38012 & & 0.21019 & 0.55684 & & 0.11804 & 0.35571 & \\
\hdashline
\multirow{2}{*}{\textbf{CineSRD (Ours)}} & AV & \colorbox{mlo}{\textbf{0.08325}} & \colorbox{mlo}{\textbf{0.41444}} & & \colorbox{mlo}{\textbf{0.10180}} & \colorbox{mlo}{\textbf{0.55392}} & & \colorbox{mlo}{\textbf{0.10272}} & \colorbox{mlo}{\textbf{0.31331}} & \\
~ & AVT & \colorbox{mlb}{\textbf{0.07561}} & \colorbox{mlb}{\textbf{0.31965}} & & \colorbox{mlb}{\textbf{0.09470}} & \colorbox{mlb}{\textbf{0.41462}} & & \colorbox{mlb}{\textbf{0.08932}} & \colorbox{mlb}{\textbf{0.29093}} & \\
\Xhline{1.0pt}
\end{tabular}
\label{tab:sdeval-main}
\end{table*}

\subsubsection{Conventional Speaker Diarization Scenario}

We validate the effectiveness of CineSRD in general speaker diarization scenarios by comparing it with other baseline methods on the public AVA-AVD \cite{modal6} benchmark. The results are shown in Table~\ref{tab:sdeval-ava}, where we report DER and Speaker Error Rate (SPKE) metrics. Since AVA-AVD is an audio-visual dataset, we eliminate the text modality in CineSRD. The results show that even without text information, CineSRD consistently outperforms all compared baselines under different clustering strategies. Compared with AHC and SC, which rely solely on traditional audio-signal clustering, AVR-Net and EC2P achieve only modest improvements through cross-modal fusion. In contrast, CineSRD achieves a more significant performance gain by integrating strategies such as visual anchor clustering and speaker turn detection. This improvement stems from the role of visual cues in correcting audio clustering errors in complex scenes. These findings indicate that CineSRD is not limited to visual media programs; it also demonstrates strong extensibility and generalization across other scenarios and open-source benchmarks.

\begin{table}[!h]
\centering
\caption{Evaluation results of AVA-AVD datasets. The performance metrics of the baseline methods are obtained from their respective papers.}
\begin{tabular}{lccc}
 \Xhline{1.0pt}
 \rowcolor{gray!20}
\textbf{Method} & \textbf{Modality} & \textbf{DER~$\downarrow$} & \textbf{SPKE~$\downarrow$} \\
\hline
AHC~\cite{ahc} & A & 0.2137 & 0.1845 \\
SC~\cite{sc} & A & 0.2131 & 0.1839 \\
AVR-Net~\cite{modal6} & AV & 0.2057 & 0.1765 \\
EC2P~\cite{aclsd} & AV & 0.2032 & 0.1740 \\
\hdashline
\textbf{CineSRD (AHC)} & AV & \colorbox{mlo}{\textbf{ 0.1951 }} & \colorbox{mlo}{\textbf{ 0.1630 }} \\
\textbf{CineSRD (SC)} & AV & \colorbox{mlb}{\textbf{ 0.1898 }} & \colorbox{mlb}{\textbf{ 0.1616 }} \\
 \Xhline{1.0pt}
\end{tabular}
\label{tab:sdeval-ava}
\end{table}

\subsection{Ablation Study}

\subsubsection{Multimodal Fusion in CineSRD}

\begin{table*}[t]
\centering
\caption{CineSRD performance across modalities using different clustering methods. The setting that uses only the textual modality refers to the pure clustering method.}
\begin{tabular}{cccc|cc|cc}
\Xhline{1.0pt}
\rowcolor{gray!20}
~ & ~ & \multicolumn{2}{c}{\textbf{Chinese}} & \multicolumn{2}{c}{\textbf{Chinese-Hard}} & \multicolumn{2}{c}{\textbf{English}} \\
\cline{3-8}
\rowcolor{gray!20}
\multirow{-2}{*}{\textbf{Clustering}} & \multirow{-2}{*}{\textbf{Modality}} 
& \textbf{DER~$\downarrow$} & \textbf{JER~$\downarrow$} & \textbf{DER~$\downarrow$} & \textbf{JER~$\downarrow$} & \textbf{DER~$\downarrow$} & \textbf{JER~$\downarrow$} \\
\hline
\multirow{3}{*}{AHC} & A & 0.13982 & 0.45223 & 0.20679 & 0.57714 & 0.12481 & 0.41017 \\
~ & AV & 0.08528 & 0.41847 & 0.10585 & 0.55233 & 0.10554 & 0.31771 \\
~ & AVT & \colorbox{mlo}{\textbf{0.07700}} & \colorbox{mlo}{\textbf{0.32119}} & \colorbox{mlo}{\textbf{0.09928}} & \colorbox{mlb}{\textbf{0.41314}} & 0.10357 & \colorbox{mlb}{\textbf{0.27094}} \\
\hdashline
\multirow{3}{*}{SC} & A & 0.14945 & 0.47114 & 0.22255 & 0.65194 & 0.20116 & 0.43402 \\
~ & AV & 0.08325 & 0.41444 & 0.10180 & 0.55392 & \colorbox{mlo}{\textbf{0.10272}} & 0.31331 \\
~ & AVT & \colorbox{mlb}{\textbf{0.07561}} & \colorbox{mlb}{\textbf{0.31965}} & \colorbox{mlb}{\textbf{0.09470}} & \colorbox{mlo}{\textbf{0.41462}} & \colorbox{mlb}{\textbf{0.08932}} & \colorbox{mlo}{\textbf{0.29093}} \\
\Xhline{1.0pt}
\end{tabular}
\label{tab:sdeval-multimodal}
\end{table*}

We present in Table~\ref{tab:sdeval-multimodal} the influence of different modality combinations on visual media speaker diarization performance. CineSRD is evaluated with both AHC and SC clustering strategies. The results show that integrating the visual modality into conventional audio-only clustering significantly reduces both DER and JER. When the text modality (speaker turn detection) is further introduced, the performance continues to improve, confirming the importance of textual information in semantic completion and boundary refinement. This trend is consistently observed across all three subsets of SubtitleSD. Overall, visual anchor clustering leverages face features to markedly enhance the reliability of clustering boundaries, while textual semantic cues effectively correct audio-visual diarization. In addition, off-screen speaker supplementation further mitigates annotation errors arising from audio-visual asynchrony or missing character detection. By jointly utilizing all three modalities, CineSRD achieves accurate alignment between lines and characters even in open-world environments.

\subsubsection{Audio-Textual Fusion in Speaker Turn Detection}

\begin{figure}[!h]
  \centering
  \includegraphics[width=0.48\textwidth]{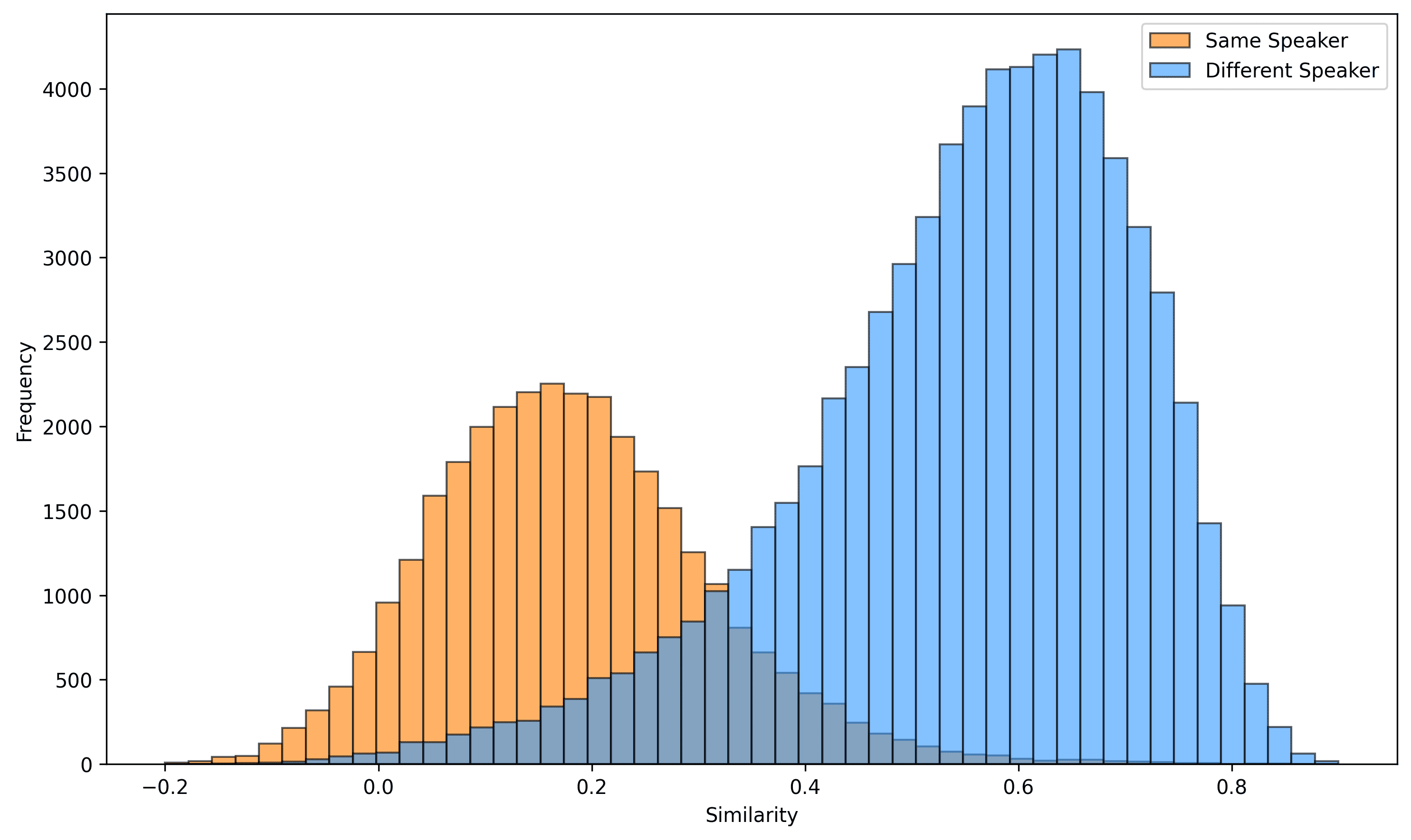}
  \caption{Label frequency distribution on Chinese subset.}
  \label{fig:stdlabel}
\end{figure}

We present the comparison of speaker turn detection performance among pure ALM (ALM Only), pure audio (Audio Only, using a similarity threshold for discrimination), and their weighted fusion (ALM + Audio) in Table~\ref{tab:turn-detection}, evaluated on the Chinese subset of SubtitleSD. The results show that although ALM also takes audio as input, using audio alone still performs better than using ALM alone, indicating that speaker turn detection mainly relies on timbre features in speech. Figure~\ref{fig:stdlabel} illustrates the similarity–frequency distribution of speaker labels between adjacent lines on the Chinese subset. The overlap between the distributions of same and different labels suggests that a portion of adjacent lines cannot be distinguished purely by timbre similarity and requires semantic understanding. In Table~\ref{tab:turn-detection}, after applying weighted fusion, both AUC and F1 scores further improve, demonstrating that the introduction of the text modality plays a crucial auxiliary role in determining whether two lines are spoken by the same speaker, thereby confirming the importance of multimodal fusion in the speaker turn detection module.

\begin{table}[!h]
\centering
\caption{Evaluation results of different methods on speaker turn detection.}
\begin{tabular}{lccc}
 \Xhline{1.0pt}
 \rowcolor{gray!20}
\textbf{Method} & \textbf{Modality} & \textbf{AUC~$\uparrow$} & \textbf{F1~$\uparrow$} \\
\hline
ALM Only & AT & 0.78974 & 0.83015 \\
Audio Only & A & 0.95862 & 0.86490 \\
ALM + Audio & AT & \colorbox{mlb}{\textbf{0.96257}} & \colorbox{mlb}{\textbf{0.92715}} \\
 \Xhline{1.0pt}
\end{tabular}
\label{tab:turn-detection}
\end{table}

\subsubsection{Weight $w$ in Speaker Turn Detection}

For the weighting hyperparameter $w$, we conduct the experiment in Figure~\ref{fig:register} to examine how the F1 metric of speaker turn detection varies with its value. The results show that speaker turn detection achieves the best performance when $w = 0.45$, and we adopt this value for all other experiments in this paper. This also indicates that in computing $P_{\text{std}}$, $S_{\text{tim}}$ carries a higher weight than $P_{\text{alm}}$, further confirming that speaker turn detection primarily relies on speech timbre features, with the textual modality playing a auxiliar role.

\begin{figure}[!h]
  \centering
  \includegraphics[width=0.48\textwidth]{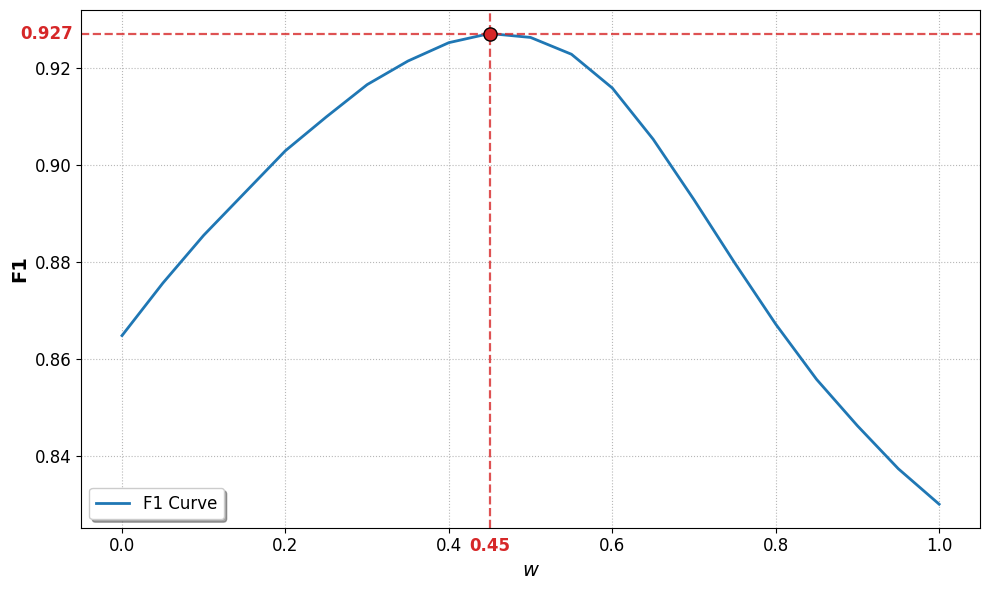}
  \caption{Impact of weight $w$ in speaker turn detection.}
  \label{fig:register}
\end{figure}

\subsubsection{Threshold $\eta$ in Speaker Supplementation}

\begin{figure}[!h]
  \centering
  \includegraphics[width=0.48\textwidth]{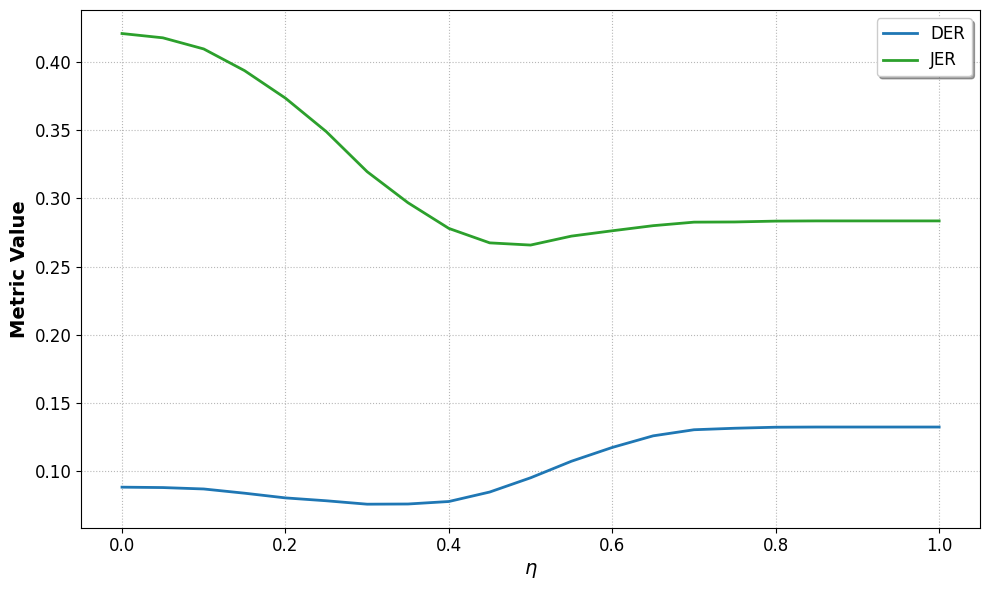}
  \caption{Impact of Threshold $\eta$ in off-screen speaker supplementation.}
  \label{fig:sdm-ct}
\end{figure}

$\eta$ serves as a threshold and is a key hyperparameter for controlling new speaker registration in the off-screen speaker supplementation strategy. To examine the influence of $\eta$ on off-screen speaker discovery, we conduct the experiment shown in Figure~\ref{fig:sdm-ct}, which illustrates the trends of DER and JER as $\eta$ varies. When $\eta$ is too small, the off-screen speaker supplementation mechanism contributes little, and most speech segments are assigned only using the existing speaker library, reducing the ability to discover new speakers. In contrast, when $\eta$ approaches 1, new speaker registration occurs frequently, which significantly increases inference latency and introduces excessive noisy speakers, degrading overall accuracy. The experimental results indicate that when $\eta$ is around 0.45, CineSRD achieves an ideal trade-off between recognition accuracy and inference efficiency, where both DER and JER reach favorable values, balancing new character discovery with controlled misassignment rates.

\section{Limitations}

Although CineSRD demonstrates strong performance, it remains a multi-stage integrated framework rather than a fully end-to-end approach. Given the current limitations of multimodal large models in handling long-form videos and their high computational costs, developing an end-to-end speaker diarization method is highly challenging. Moreover, since CineSRD relies on face detection and embedding models trained on real human face data, its effectiveness is limited when applied to programs of genres such as cartoon or animation. For our SubtitleSD benchmark, while the Chinese subset is sufficiently large, the English subset remains relatively small, which restricts its utility for further improving speaker diarization performance through training on the English subset.

\section{Conclusion}

In this work, we investigated speaker diarization in the open-world domain of visual media, where long-form narratives, large speaker sets, cross-modal asynchrony, and in-the-wild variability present significant challenges. To address these issues, we introduced CineSRD, a unified multimodal framework that leverages visual anchor clustering, speaker turn detection, and off-screen speaker supplementation to robustly associate each line with its corresponding character. We further constructed SubtitleSD, a dedicated bilingual benchmark spanning diverse program genres. Comprehensive experiments demonstrate that CineSRD achieves superior performance on SubtitleSD and maintains strong generalization in conventional diarization scenarios, validating its effectiveness and robustness in complex audiovisual environments.

\section*{Acknowledgments}

The authors would like to thank all the anonymous reviewers for their helpful and insightful comments.

{
    \small
    \bibliographystyle{ieeenat_fullname}
    \bibliography{main}

@String(ICASSP=	{ICASSP})

@String(AAAI = {AAAI})

@inproceedings{vd1,
  title={From Speech-to-Speech Translation to Automatic Dubbing},
  author={Federico, Marcello and Enyedi, Robert and Barra-Chicote, Roberto and Giri, Ritwik and Isik, Umut and Krishnaswamy, Arvindh and Sawaf, Hassan},
  booktitle={Proceedings of the 17th International Conference on Spoken Language Translation},
  pages={257--264},
  year={2020}
}

@inproceedings{vd2,
  title={Videodubber: Machine translation with speech-aware length control for video dubbing},
  author={Wu, Yihan and Guo, Junliang and Tan, Xu and Zhang, Chen and Li, Bohan and Song, Ruihua and He, Lei and Zhao, Sheng and Menezes, Arul and Bian, Jiang},
  booktitle={Proceedings of the AAAI Conference on Artificial Intelligence},
  volume={37},
  number={11},
  pages={13772--13779},
  year={2023}
}

@inproceedings{vd3,
  title={Isometric Neural Machine Translation using Phoneme Count Ratio Reward-based Reinforcement Learning},
  author={Mhaskar, Shivam and Shah, Nirmesh and Zaki, Mohammadi and Gudmalwar, Ashishkumar and Wasnik, Pankaj and Shah, Rajiv},
  booktitle={Findings of the Association for Computational Linguistics: NAACL 2024},
  pages={3966--3976},
  year={2024}
}

@inproceedings{aclsd,
  title={Integrating Audio, Visual, and Semantic Information for Enhanced Multimodal Speaker Diarization on Multi-party Conversation},
  author={Cheng, Luyao and Wang, Hui and Deng, Chong and Zheng, Siqi and Chen, Yafeng and Huang, Rongjie and Zhang, Qinglin and Chen, Qian and Li, Xihao and Wang, Wen},
  booktitle={Proceedings of the 63rd Annual Meeting of the Association for Computational Linguistics (Volume 1: Long Papers)},
  pages={19914--19928},
  year={2025}
}

@inproceedings{talknet,
  title={Is someone speaking? exploring long-term temporal features for audio-visual active speaker detection},
  author={Tao, Ruijie and Pan, Zexu and Das, Rohan Kumar and Qian, Xinyuan and Shou, Mike Zheng and Li, Haizhou},
  booktitle={Proceedings of the 29th ACM international conference on multimedia},
  pages={3927--3935},
  year={2021}
}

@article{sc,
  title={A tutorial on spectral clustering},
  author={Von Luxburg, Ulrike},
  journal={Statistics and computing},
  volume={17},
  number={4},
  pages={395--416},
  year={2007},
  publisher={Springer}
}

@inproceedings{eres2netv2,
  title={ERes2NetV2: Boosting Short-Duration Speaker Verification Performance with Computational Efficiency},
  author={Chen, Yafeng and Zheng, Siqi and Wang, Hui and Cheng, Luyao and Chen, Qian and Zhang, Shiliang and Li, Junjie},
  booktitle={Proc. Interspeech 2024},
  pages={3245--3249},
  year={2024}
}

@inproceedings{retinaface,
  title={Retinaface: Single-shot multi-level face localisation in the wild},
  author={Deng, Jiankang and Guo, Jia and Ververas, Evangelos and Kotsia, Irene and Zafeiriou, Stefanos},
  booktitle={Proceedings of the IEEE/CVF conference on computer vision and pattern recognition},
  pages={5203--5212},
  year={2020}
}

@inproceedings{curricular,
  title={Curricularface: adaptive curriculum learning loss for deep face recognition},
  author={Huang, Yuge and Wang, Yuhan and Tai, Ying and Liu, Xiaoming and Shen, Pengcheng and Li, Shaoxin and Li, Jilin and Huang, Feiyue},
  booktitle={proceedings of the IEEE/CVF conference on computer vision and pattern recognition},
  pages={5901--5910},
  year={2020}
}

@inproceedings{app1,
  title={Addressee and response selection for multi-party conversation},
  author={Ouchi, Hiroki and Tsuboi, Yuta},
  booktitle={Proceedings of the 2016 Conference on Empirical Methods in Natural Language Processing},
  pages={2133--2143},
  year={2016}
}

@article{app2,
  title={MPC-BERT: A pre-trained language model for multi-party conversation understanding},
  author={Gu, Jia-Chen and Tao, Chongyang and Ling, Zhen-Hua and Xu, Can and Geng, Xiubo and Jiang, Daxin},
  journal={arXiv preprint arXiv:2106.01541},
  year={2021}
}

@inproceedings{app3,
  title={Speaker diarization with LSTM},
  author={Wang, Quan and Downey, Carlton and Wan, Li and Mansfield, Philip Andrew and Moreno, Ignacio Lopz},
  booktitle={2018 IEEE International conference on acoustics, speech and signal processing (ICASSP)},
  pages={5239--5243},
  year={2018},
  organization={IEEE}
}

@inproceedings{app4,
  title={Fully supervised speaker diarization},
  author={Zhang, Aonan and Wang, Quan and Zhu, Zhenyao and Paisley, John and Wang, Chong},
  booktitle={ICASSP 2019-2019 IEEE International Conference on Acoustics, Speech and Signal Processing (ICASSP)},
  pages={6301--6305},
  year={2019},
  organization={IEEE}
}

@article{app5,
  title={An overview of automatic speaker diarization systems},
  author={Tranter, Sue E and Reynolds, Douglas A},
  journal={IEEE Transactions on audio, speech, and language processing},
  volume={14},
  number={5},
  pages={1557--1565},
  year={2006},
  publisher={IEEE}
}

@article{app6,
  title={Speaker diarization: A review of recent research},
  author={Anguera, Xavier and Bozonnet, Simon and Evans, Nicholas and Fredouille, Corinne and Friedland, Gerald and Vinyals, Oriol},
  journal={IEEE Transactions on audio, speech, and language processing},
  volume={20},
  number={2},
  pages={356--370},
  year={2012},
  publisher={IEEE}
}

@inproceedings{app7,
  title={Approaches and applications of audio diarization},
  author={Reynolds, Douglas A and Torres-Carrasquillo, P},
  booktitle={Proceedings.(ICASSP'05). IEEE International Conference on Acoustics, Speech, and Signal Processing, 2005.},
  volume={5},
  pages={v--953},
  year={2005},
  organization={IEEE}
}

@article{app8,
  title={Auto-tuning spectral clustering for speaker diarization using normalized maximum eigengap},
  author={Park, Tae Jin and Han, Kyu J and Kumar, Manoj and Narayanan, Shrikanth},
  journal={IEEE Signal Processing Letters},
  volume={27},
  pages={381--385},
  year={2019},
  publisher={IEEE}
}

@inproceedings{app9,
  title={Fine-grained Video Dubbing Duration Alignment with Segment Supervised Preference Optimization},
  author={Cui, Chaoqun and Huang, Liangbin and Wang, Shijing and Tong, Zhe and Huang, Zhaolong and Zeng, Xiao and Liu, Xiaofeng},
  booktitle={Proceedings of the 63rd Annual Meeting of the Association for Computational Linguistics (Volume 1: Long Papers)},
  pages={4524--4546},
  year={2025}
}

@inproceedings{modal1,
  title={Speaker diarization with PLDA i-vector scoring and unsupervised calibration},
  author={Sell, Gregory and Garcia-Romero, Daniel},
  booktitle={2014 IEEE Spoken Language Technology Workshop (SLT)},
  pages={413--417},
  year={2014},
  organization={IEEE}
}

@article{modal3,
  title={Bayesian hmm clustering of x-vector sequences (vbx) in speaker diarization: theory, implementation and analysis on standard tasks},
  author={Landini, Federico and Profant, J{\'a}n and Diez, Mireia and Burget, Luk{\'a}{\v{s}}},
  journal={Computer Speech \& Language},
  volume={71},
  pages={101254},
  year={2022},
  publisher={Elsevier}
}

@inproceedings{modal4,
  title={Who said that?: Audio-visual speaker diarisation of real-world meetings},
  author={Chung, Joon Son and Lee, Bong Jin and Han, Icksang},
  booktitle={Proceedings of the Annual Conference of the International Speech Communication Association, INTERSPEECH},
  volume={2019},
  pages={371--375},
  year={2019}
}

@article{modal5,
  title={Audio-visual speaker diarization based on spatiotemporal bayesian fusion},
  author={Gebru, Israel D and Ba, Sileye and Li, Xiaofei and Horaud, Radu},
  journal={IEEE transactions on pattern analysis and machine intelligence},
  volume={40},
  number={5},
  pages={1086--1099},
  year={2017},
  publisher={IEEE}
}

@inproceedings{modal6,
  title={Ava-avd: Audio-visual speaker diarization in the wild},
  author={Xu, Eric Zhongcong and Song, Zeyang and Tsutsui, Satoshi and Feng, Chao and Ye, Mang and Shou, Mike Zheng},
  booktitle={Proceedings of the 30th ACM International Conference on Multimedia},
  pages={3838--3847},
  year={2022}
}

@article{modal7,
  title={Multimodal clustering with role induced constraints for speaker diarization},
  author={Flemotomos, Nikolaos and Narayanan, Shrikanth},
  journal={arXiv preprint arXiv:2204.00657},
  year={2022}
}

@article{modal8,
  title={Multimodal speaker segmentation and diarization using lexical and acoustic cues via sequence to sequence neural networks},
  author={Park, Tae Jin and Georgiou, Panayiotis},
  journal={arXiv preprint arXiv:1805.10731},
  year={2018}
}

@inproceedings{modal9,
  title={Bertraffic: Bert-based joint speaker role and speaker change detection for air traffic control communications},
  author={Zuluaga-Gomez, Juan and Sarfjoo, Seyyed Saeed and Prasad, Amrutha and Nigmatulina, Iuliia and Motlicek, Petr and Ondrej, Karel and Ohneiser, Oliver and Helmke, Hartmut},
  booktitle={2022 IEEE Spoken Language Technology Workshop (SLT)},
  pages={633--640},
  year={2023},
  organization={IEEE}
}

@inproceedings{modal10,
  title={Exploring Speaker-Related Information in Spoken Language Understanding for Better Speaker Diarization},
  author={Cheng, Luyao and Zheng, Siqi and Qinglin, Zhang and Wang, Hui and Chen, Yafeng and Chen, Qian},
  booktitle={Findings of the Association for Computational Linguistics: ACL 2023},
  pages={14068--14077},
  year={2023}
}

@article{modal11,
  title={Improving speaker diarization using semantic information: Joint pairwise constraints propagation},
  author={Cheng, Luyao and Zheng, Siqi and Zhang, Qinglin and Wang, Hui and Chen, Yafeng and Chen, Qian and Zhang, Shiliang},
  journal={arXiv preprint arXiv:2309.10456},
  year={2023}
}

@article{intro1,
  title={Acoustic beamforming for speaker diarization of meetings},
  author={Anguera, Xavier and Wooters, Chuck and Hernando, Javier},
  journal={IEEE Transactions on Audio, Speech, and Language Processing},
  volume={15},
  number={7},
  pages={2011--2022},
  year={2007},
  publisher={IEEE}
}

@inproceedings{intro2,
  title={A real-time speaker diarization system based on spatial spectrum},
  author={Zheng, Siqi and Huang, Weilong and Wang, Xianliang and Suo, Hongbin and Feng, Jinwei and Yan, Zhijie},
  booktitle={ICASSP 2021-2021 IEEE International Conference on Acoustics, Speech and Signal Processing (ICASSP)},
  pages={7208--7212},
  year={2021},
  organization={IEEE}
}

@article{ahc, title={Efficient agglomerative hierarchical clustering}, author={Bouguettaya, Athman and Yu, Qi and Liu, Xumin and Zhou, Xiangmin and Song, Andy}, journal={Expert Systems with Applications}, volume={42}, number={5}, pages={2785--2797}, year={2015}, publisher={Elsevier} }

@article{Qwen2-Audio, title={Qwen2-audio technical report}, author={Chu, Yunfei and Xu, Jin and Yang, Qian and Wei, Haojie and Wei, Xipin and Guo, Zhifang and Leng, Yichong and Lv, Yuanjun and He, Jinzheng and Lin, Junyang and others}, journal={arXiv preprint arXiv:2407.10759}, year={2024} }

@inproceedings{der,
  title={The rich transcription 2006 spring meeting recognition evaluation},
  author={Fiscus, Jonathan G and Ajot, Jerome and Michel, Martial and Garofolo, John S},
  booktitle={International Workshop on Machine Learning for Multimodal Interaction},
  pages={309--322},
  year={2006},
  organization={Springer}
}

@inproceedings{jer,
  title={The Second DIHARD Diarization Challenge: Dataset, Task, and Baselines},
  author={Ryant, Neville and Church, Kenneth and Cieri, Christopher and Cristia, Alejandrina and Du, Jun and Ganapathy, Sriram and Liberman, Mark},
  booktitle={Interspeech 2019},
  pages={978--982},
  year={2019}
}

@inproceedings{AMI,
  title={The AMI meeting corpus: A pre-announcement},
  author={Carletta, Jean and Ashby, Simone and Bourban, Sebastien and Flynn, Mike and Guillemot, Mael and Hain, Thomas and Kadlec, Jaroslav and Karaiskos, Vasilis and Kraaij, Wessel and Kronenthal, Melissa and others},
  booktitle={International workshop on machine learning for multimodal interaction},
  pages={28--39},
  year={2005},
  organization={Springer}
}

@inproceedings{VoxConverse,
  title     = {Spot the Conversation: Speaker Diarisation in the Wild},
  author    = {Joon Son Chung and Jaesung Huh and Arsha Nagrani and Triantafyllos Afouras and Andrew Zisserman},
  year      = {2020},
  booktitle = {Interspeech 2020},
  pages     = {299--303},
  doi       = {10.21437/Interspeech.2020-2337},
  issn      = {2958-1796},
}

@article{hermes,
  title={Hermes the Polyglot: A Unified Framework to Enhance Expressiveness for Multimodal Interlingual Subtitling},
  author={Cui, Chaoqun and Wang, Shijing and Huang, Liangbin and Gu, Qingqing and Huang, Zhaolong and Zeng, Xiao and Mao, Wenji},
  journal={arXiv preprint arXiv:2602.00597},
  year={2026}
}
}


\end{document}